# Developing a Gender Classification Approach in Human Face Images using Modified Local Binary Patterns and Tani-Moto based Nearest Neighbor Algorithm

Shervan Fekri-Ershad[1,2,*]

[1] Faculty of computer engineering, Najafabad branch, Islamic azad university, Najafabad, Iran

[2] Big data research center, Najafabad branch, Islamic azad university, Najafabad, Iran

Email: fekriershad@pco.iaun.ac.ir

*Abstract* – **Human identification is a much attention problem in computer vision. Gender classification plays an important role in human identification as preprocess step. So far, various methods have been proposed to solve this problem. Absolutely, classification accuracy is the main challenge for researchers in gender classification. But, some challenges such as rotation, gray scale variations, pose, illumination changes may be occurred in smart phone image capturing. In this respect, a multi step approach is proposed in this paper to classify genders in human face images based on improved local binary patters (MLBP). LBP is a texture descriptor, which extract local contrast and local spatial structure information. Some issues such as noise sensitivity, rotation sensitivity and low discriminative features can be considered as disadvantages of the basic LBP. MLBP handle disadvantages using a new theory to categorize extracted binary patterns of basic LBP. The proposed approach includes two stages. First of all, a feature vector is extracted for human face images based on MLBP. Next, non linear classifiers can be used to classify gender. In this paper nearest neighborhood classifier is evaluated based on Tani-Moto metric as distance measure. In the result part, two databases, self-collected and ICPR are used as human face database. Results are compared by some state-of-the-art algorithms in this literature that shows the high quality of the proposed approach in terms of accuracy rate. Some of other main advantages of the proposed approach are rotation invariant, low noise sensitivity, size invariant and low computational complexity. The proposed approach decreases the computational complexity of smartphone applications because of reducing the number of database comparisons. It can also improve performance of the synchronous applications in the smarphones because of memory and CPU usage reduction.**

## 1. INTRODUCTION

Human's face is a major feature in visual machine learning and image processing systems to identify desired aims. A face carries discriminative and separable information consisting gender, age, ethnicity, etc [1]. Face information is applicable in many cases such as human-computer interaction, image retrieval, biometric authentication, drivers monitoring, human-robot interaction, sport competition senses analysis, video summarizing, and image/video indexing.

Gender classification may decrease the computational complexity of human identification systems. In near all of the biometric identification systems, a big database is used includes large number of humans, whether man or woman. Many smartphone applications usually use the main memory of the mobile to perform the identification. Therefore, reducing the computational complexity of this task will greatly affect the performance of smartphones.

Suppose that in a big database, there are *n* images from human faces. In a human identification system, the query image should be compared to all of the images inside the database, so that the diagnosis can be done accurately. It means, *n* times the feature comparison will be needed. Now, if the gender of the query image is clear, it's just that, according to the likelihood rule, the features of the query image will be compared to 50 percent of the images inside the database (n/2 samples). Therefore, if the computational complexity of the gender classification is less than the computational of comparing the query image with 50% of the base images, the act of gender classification can greatly increase the performance of the smart phones. This claim shows that, unlike to belief of many researchers, gender classification can be one of the most widely used areas in image processing, with the application in new technologies such as smart phones.

On the other hand, in some of the smartphone applications, gender classification plays an important role. For example in some cases, it is important for parents to identify the gender of their child's friend. The process of a human identification system is shown in Fig. 1 based on two different ways in smartphones. As mentioned above, the proposed method can improve mobile performance in either of the two modes shown in Figure 1.

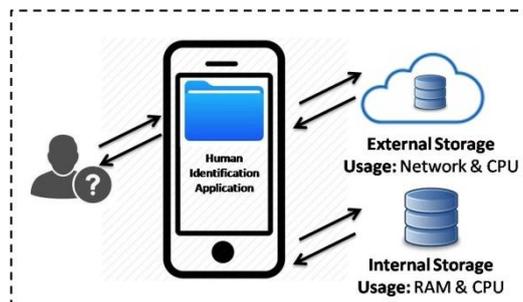

Fig. 1. The relation between smart phone and properties of human identification applications

In this regard, the main goal of this paper is to prepare a robust approach for gender classification from human faces. Almost all of the works in gender classification involves two steps: extracting features from faces and classifying those features using labeled data. They mostly differ in the way these two steps are performed. Therefore gender classification approaches can be categorized based on the feature extraction and classification methods. Usually two strategies are followed for feature extraction as, appearance-based and geometry-based features. Near all of the appearance-based methods considering the whole image rather than local features that are present in different parts of the face. On the other hand in geometry-based approaches, the geometric features such as distance between eyes, face width, length, thickness of nose and etc, of a face are considered. The proposed approach extracts local texture information, so it can be categorized as appearance-based approaches. Hence, in the next section, only the works related to appearance-based methods are discussed. For the classification phase, previous works usually used support vector machine, k- nearest neighbors, neural networks, distribution modeling, etc.

Local binary pattern (LBP) is a non-parametric operator which describes the local contrast and local spatial structure of an image. LBP commonly is used in computer vision systems to describe image texture. It provides discriminative texture features to identify image content. In this paper an approach is proposed for gender classification of human based on improved version of LBP. First of all, improved LBP is used to describe input human facial image as a feature vector. Next, a classification phase is needed to identify the gender of the test image. In this paper, K-nearest neighborhood classifier is evaluated based on Tani-Moto metric, which is notified by E-KNN in this paper.

This paper is organized as follows: Section 2 is about the related works. Section 3 is discussed the proposed approach in two steps, feature extraction (Sub-Section 3.1.) and Proposed Classification (Sub-Section 3.2.). Experimentally results are presented in section4. The proposed approach is compared with other methods in terms of rotation invariant, computational complexity, size invariance, noise sensitivity in sub-section 4.1. Finally conclusion is included.

## 2. RELATED WORKS

As it was mentioned in previous section, gender classification methods are different in feature extraction and classification algorithms. As a recent work, a multi agent system is proposed in [2] to classify gender and age in human face images. An innovative method is proposed in [2] for image acquisition, preprocessing and final classification. Experimental results in [2] shows that Fisherfaces method is more effective than multi layer perception (MLP). Also, its training time is shorter because of dimension reducing.

Scalzo et al., in [6], present a hierarchical feature fusion model for gender classification. The proposed model in [6] has the ability to combine local patches determined by evolutionary algorithm. The novelty point of [6] is to use a two-level framework, which combines feature fusion and decision fusion into a unified model. The justification of [6] is not claiming its performance for challengeable gender datasets. In [7], non-linear support vector machine (SVM) is used to classify faces from thumbnail images with low resolutions. In [7], SVM is set up with Gaussian RBF and cubic polynomial kernels. Moghaddam et al. [7], also experimented with other types of classifiers including different types of linear discrimination analysis (LDA), RBFs, K-Nearest Neighbor. They used a database with 1,755 thumbnail images including 1044 males and 711 females. Best accuracy was obtained using SVM along with Gaussian RBF kernel with error rate of 3.38 percent for males and 2.05 percent for females. Regardless of high detection rate, their method [7] has some disadvantages such as: collecting a database of thumbnails is not possible in many case, the computational complexity of [7] was very high because of evaluation and using RBF kernel.

Shakhnarovich, et al. [9] applied AdaBoost to the features used by the face detection system which created on 24×24 pixel images collected by crawling the web. They obtained an accuracy of 79 percent. The performance of some well-know human face recognition algorithms are compared in [11]. In [12], S. Buchala proposed a gender classification system based on a sub set of selected features using principal component analysis (PCA), Self organizing maps (SOM), and curvilinear component analysis (CCA). Best results are obtained using 759 PCA components, about 92.25 percent detection rate. Main disadvantage of [12] is over-fitting because, the size of the faces is restricted to 128x128 pixels and the test set is composed by only little number of faces. In [13], an Adaboost system is proposed for gender classification with manually aligned faces. They evaluated a database of human faces by varying face image scaling, translation, and rotation. An experimental comparison between the Adaboost and an SVM classifier are proposed in [13] too.

Most of the previous works use manual detection and alignment of the face images which has been shown to improve performance [1, 14]. Only few studies have automatically extracted faces using a face detector [5, 6, 9]. Some major challenges of gender classification are to account for the effects of illumination, background clutter, pose and etc. Practical systems have to be robust enough to take these issues into consideration. Most of the work in gender classification assumes that the frontal views of faces, which are pre-aligned and free of distracting background clutters, are available [2].

In [3], an approach is proposed for gender classification based on two steps: feature extraction and classification. First of all, the input image is converted to HSV color space. Next, adaptive feature set is extracted for localize face images. Finally, SVM is used to classify two gender groups. Good localization for face detecting is one of the main advantages of [3]. But, high computational complexity can be considered as one of the main disadvantages of [3] which reduce its usability in online applications.

Recently, an image processing descriptor called Local Binary Patterns (LBP) first has been proposed by ojala et al., in [15], to analysis textures. LBP has been used in various vision applications such as face recognition [16], texture classification [17], surface defect detection [18, 19], image retrieval [20] and etc. In [21], the authors used local binary patterns as feature descriptor for gender classification. The authors extracted histogram based on primary version of LBP. Lian and Liu [22] used support vector machine as classifier correlated with local binary patterns.

The primary version of local binary patters which is used in [21, 22] as feature extractor, is sensitive to zoom, rotate, gray-scale and size. In this respect in [17] modified two dimensional version of local binary patters are proposed and its high quality to analysis textures are discussed. The proposed feature extraction algorithm in this paper is based on modified version of local binary patterns. In this respect, proposed approach is categorized in appearance based methods. Also, the nearest neighbor is applied as classifier, which is evaluated using Tani-Moto metric as distance metric. In the result part, the detection rate of the proposed approach is compared with some well know methods in this literature. The results show the high quality of the proposed to classify genders accurately. Rotation invariant, gray-scale invariant, size invariant, low sensitivity to noise, and low computational complexity are some other advantages of the proposed approach. All of the advantage are discussed in a clear way separately.

### 3. PROPOSED GENDER CLASSIFICATION APPROACH

Gender classification is categorized as a visual pattern classification problem. Near all of the works in gender classification involves two main steps: a) extracting features from faces b) classifying those features using labeled data. They mostly differ in the way these two steps are performed. The proposed approach has a main flowchart as shown in figure 2. As it is shown in the figure 2, the novelty of the proposed approach is providing local texture information as features using MLBP. Also, the K-nearest neighborhood is evaluated using Tani-Moto metric to improve the recognition rate.

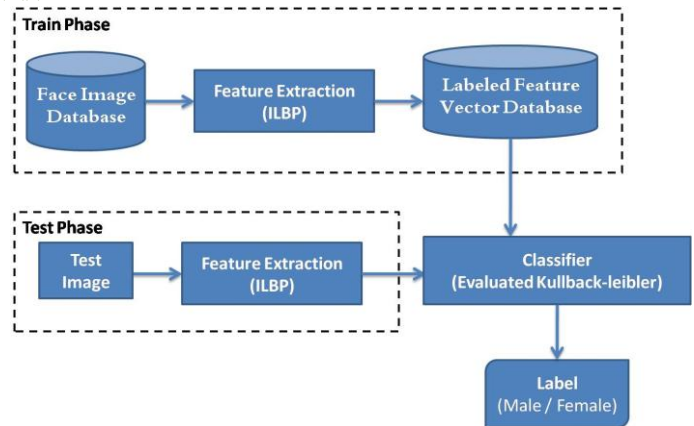

Fig. 2. The flowchart of the proposed gender classification system

### 3.1. Feature extraction step

Detection rate is much related to the discrimination of extracted features. The main visual difference between the human face images of man and female is usually in the face texture. For example, the body texture of men is usually thicker and has more edges than women. The results of the researchers in previous papers have shown that color and shape information cannot make a significant difference between these two groups. So, in this paper texture information is extracted from human faces using modified local binary patterns (MLBP).

### 3.1.1. Basic local binary patterns

In [15], the authors introduced an image descriptor titled local binary pattern to consider texture information. The LBP is an efficient operator which describes the local contrast and local spatial structure of an image in multi resolutions. Experimental results in [15], showed high severability and discriminative power of LBP for texture classification. In order to evaluate the LBP, at a given pixel position ($x_c$, $y_c$), LBP is defined as an ordered set of binary intensity comparisons between the center and its surrounding neighbors. In mostly cases, neighborhoods are considered circular because of achieving the rotation invariant. Points which the coordination's are not exactly located at the center of pixel would be found by bilinear interpolation technique. In Figure 3, some circular neighbors are shown with different radius from R=1 to R=2 and respected different number of neighborhoods pixels (P). Of course in some cases such as surface defect detection [18, 19] the squared neighborhood is considered to reduce computational complexity.

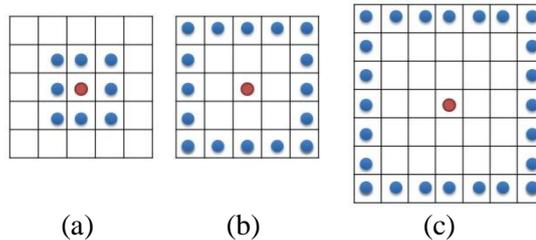

(a)　　　　(b)　　　　(c)

Fig. 3. Some examples of squared neighborhoods in LBP process
(a) P=8, R=1 (b) P=16, R=2 (c) P=24, R=3

Now, the LBP are defined at desired neighborhood using Eq. (1).

$$LBP_{P,R}(X_c, Y_c) = \sum_{k=1}^{P} \omega(g_k - g_c) 2^{k-1} \qquad (1)$$

Where, "$g_c$" corresponds to the grey value of the center pixel and "$g_k$" to the grey values of the $k^{th}$ neighborhood pixel. So, P will be the total number of neighborhoods in desired manner, and function $\Omega(x)$ is defined as follows:

$$\omega(x) = \begin{cases} 1 & if \quad x \geq 0 \\ 0 & else \end{cases} \qquad (2)$$

The $LBP_{P,R}$ operator produces ($2^P$) different output values, corresponding to the $2^P$ different binary patterns that can be formed by the P pixels in the neighbor set and radius R.

### 3.1.2. Modified local binary patterns

Low discrimination and high computational complexity are disadvantage of basic LBP which are discussed with details in [15] by Ojala et al. To solve these problems, in [17] the authors defined a uniformity measure "U" to categorize extracted binary patterns in significant groups. U, corresponds to the number of spatial transitions (bitwise 0/1 changes) in the extracted binary pattern. The computing process of uniformity measure is shown in Eq. (3). For example, patterns 01001100 have U value of 4, while 11000001 have U value of 2.

$$U\left(LBP_{P,R}(x_c, y_c)\right) = |\omega(g_1 - g_c) - \omega(g_P - g_c)| + \sum_{k=2}^{P} \omega(g_k - g_c) - \omega(g_{k-1} - g_c) \qquad (3)$$

Next, Patterns are grouped in two classes. Hence, patterns with uniformity amount less than $U_T$ are categorized as uniform patterns. The patterns with uniformity more than $U_T$ classified as non-uniforms. Finally, a label is assigned to each neighborhood using Eq. (4). The total number of ones in extracted binary patterns is considered as label for uniform patterns, and label P+1 is assigned to all of the non uniform patterns.

$$MLBP_{P,R}^{riu_T} = \begin{cases} \sum_{k=1}^{P}(g_k - g_c) & if \quad U(LBP_{P,R}) \leq U_T \\ P+1 & elsewhere \end{cases} \qquad (4)$$

In notation of *MLBP* operator, the "$riu_T$" reflects using "uniform" patterns that have U value of at most $U_T$. According to Eq. (4), applying $LBP_{P,R}$ will assign a label from "0" to "P" to uniform patterns and label "P+1" to non-uniform patterns. In using $MLBP_{P,R}$ just one label (P+1) is assigned to all of the non-uniform patterns. In this respect, $U_T$ should be optimized that uniform labels cover majority patterns in the image. Experimental results in [18, 19], show that if $U_T$ is selected equal to (P/4), only a negligible portion of the patterns in the texture takes label P+1.

As it was described, a label is assigned to each neighborhood. Regarding the Eq. (4), if the number of neighbors is considered "P" pixels, applying $LBP_{P,R}$ will assign a label from zero to "P" to uniform segments and label "P+1" to non-uniform segments. Using uniformity process make MLBP resistant to image rotation and gray scale variations. Image rotation may be disturb the position of neighbors in relation with center pixel, but the relation between the neighbors will be same after rotation. A visual example is shown in figure 4.

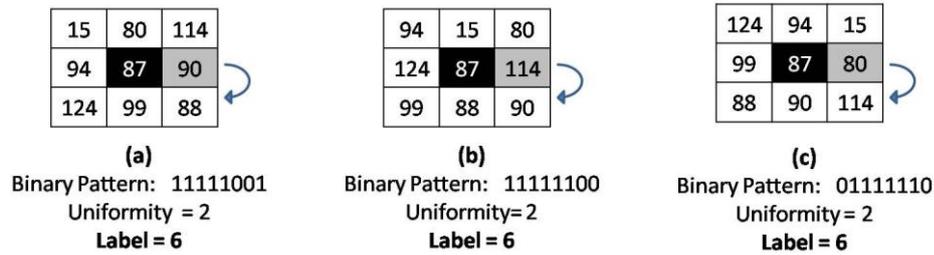

Fig. 4. An example of rotation effect on MLBP process

Applying MLBP extract local texture information based on some predefined label. In order to make it ready for classification, texture information should be converted to a numeric feature vector. So, for every image, a feature vector can be extracted with "P+2" dimensions. In this respect, first $MLBP_{P,R}$ should applied on the whole image and the labels are assigned to neighbors. Then the occurrence probability of each label in the image is regarded as one of the dimensions of the feature vector. The occurrence probability of a specific label in the image can be approximated by the ratio of the number of that label to the number of all labels (Eq. (5)).

$$F = \left\{\frac{f_0}{N_{total}}, \frac{f_1}{N_{total}}, \frac{f_2}{N_{total}}, \dots, \frac{f_{p+1}}{N_{total}}\right\} \quad (5)$$

$f_i$ is the number of neighbors with label $i$, and $N_{total}$ is the total number of neighbors. The extracted feature vector ($F$) is probabilistic which means the addition of all dimension values are one.

### 3.1.3. Image Preprocessing

Feature extraction descriptors play important role in all of the computer vision applications. So, this step should be done in a specific situation to provide maximum accuracy. In this respect, preprocessing is a necessary sub step here. We do preprocessing with the aim of increase contrast quality and normalize the output features. So, first a Gaussian filter is used to increase noise of input images. Next, all of the dataset images are resized to the same size W×W to achieve normalized feature vectors.

### 3.2. Classification Phase

K-nearest neighborhood (KNN) is a simple and non-linear classifier which is used in cases that class samples are disturbed in problem space. It is a lazy classifier that stores all available cases and classifies new cases based on a similarity measure (e.g. distance functions). A case is classified by a majority vote of its neighbors, with the case being assigned to the class most common among its T nearest samples measured by a distance function. Many different similarity or distance measures can be used to handle it. Related researches show that different similarity/distance measures may effect on performance of the systems.

For instance, in [34], the authors used cosine as a distance measure, for face verification. Pearson correlation similarity is proposed in [35] for text classification, which provided good accuracy. In information theory, the Tani-Moto metric is a symmetric distance measure that considers the probability of maximum difference between two samples in all dimensions. In [23], the authors listed more than 50 different similarity and distance metrics in terms of binary, nominal, numeric, etc.

One of main novelties of the proposed approach is using evaluated K-nearest neighborhood as classifier. In this regards, near all of the similarity/distance criteria in [23] was performed in KL divergence. Finally, Tani-Moto metric provided highest classification accuracy rate. The extracted vector using MLBP is a probabilistic measure, where the sum of all dimension values is one. The Tani-Moto metric is specialized for probabilistic vectors. Tani-Moto distance between two probabilistic feature vectors like *A* and *B*, is computed as described in the Eq. (6).

$$D_{TM}(A,B) = \sum_{k=0}^{P+1}(\max(A_k,B_k) - \min(A_k,B_k))/\sum_{k=0}^{P+1}\max(A_k,B_k) \quad (6)$$

Where, $D_{TM}$ is the Tani-Moto distance and A shows the feature vector which is extracted for train image A. Also, $A_k$ shows the value of $k^{th}$ dimension in vector A. B shows the feature vector which is extracted for test image B. Also, $B_k$ shows the value of $k^{th}$ dimension in vector B. The Tani-Moto is a distance measure, since minimization of $D_{TM}$ shows the similarity to specific class. In our application, *A* and *B* can be considered as extracted feature vectors for two texture images using Eq. (6). In our experiments, K nearest samples to the test image is selected based on Tani-Moto distance. Finally, test image is classified by a majority vote of its T nearest neighbors. It is denoted by E-KNN in this paper as follows.

## 4. EXPERIMENTAL RESULTS

The main of this paper is to propose a novel approach for gender classification in human face images. In this respect, ICPR database is used as a benchmark human face image to evaluate the performance of the proposed approach. Also, a self collected dataset is provided of human face images included popular challenges such as rotation, pose, scale variation, etc.

In order to evaluate the performance, 200 human face images are selected from wiki dataset [38], in different pose, face orientation, zoom and size. The collected database includes 100 face images of females and 100 images for male. The proposed gender classification system is applied on total database images. Finally, the evaluated T-KL which is discussed in section 3.2 was used to classify genders (Two Labels=male & female) based on T=1, 3, 5. The classification accuracy rate is shown in table 1. Hence, some other well-know classifiers such as SVM, Simple KNN, J48 Tree and Naïve Bayes are used for classification as follows. in all cases, 10-fold method is performed as validation method.

**Table1.** Proposed gender classification accuracy rate (%) using different classifiers on self-collected database

| Classifier | Acc (%) | Classifier | Acc (%) |
|---|---|---|---|
| E-1NN | 93.24 | 1-NN | 88.04 |
| E-3NN | **94.17** | 3-NN | 90.25 |
| E-5NN | 93.02 | 5-NN | 86.52 |
| SVM | 90.83 | J48 Tree | 84.22 |
| Naïve Bayes | 79.54 | MLP | 94.14 |

The maximum classification rate is provided using evaluated T-KL with T=3. In table2, the proposed gender classification approach is compared with some other state-of-the-art methods in this literature. In [22], an approach is proposed based on primary version of LBP and SVM as classifier. In [9], Shakhnarovich et al. used adaboost to propose fast and ethnicity gender classification for real time face images. In [24], an appearance-based method is proposed based on Gabor features and Fuzzy SVM as classifier. Some of the self-collected database images are shown in Figure 5.

**Table2.** Proposed gender classification accuracy rate (%) in comparison with well-known approaches on self collected Database

| Method | Classification Accuracy (%) |
|---|---|
| **Proposed Approach** | **94.17** |
| LBP + SVM [22] | 91.56 |
| Statistical Features + Adaboost [9] | 82.45 |
| LBPH+ Statistical Features + Adaboost/SVM [25] | 90.83 |
| Patch Based + Bayesian [26] | 89.35 |

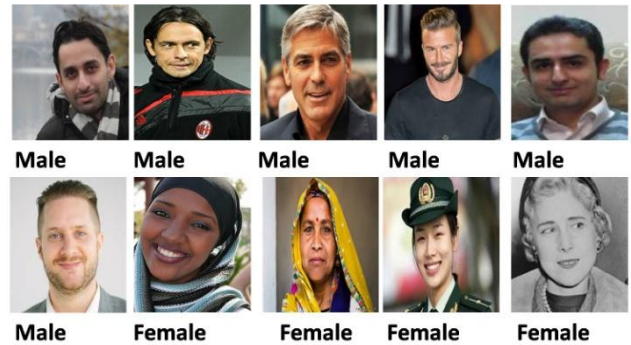

Fig. 5. Some of the self-collected dataset

Gender classification near all of the times is performed on human face images. There are many benchmarks to evaluate the performance of face recognition, face detection, gender classification or facial identification methods. ICPR is a state-of-the-art benchmark for face image analysis [28]. Different pose, skin color, illumination variation are some of the ICPR characteristics. In this respect, our proposed approach is applied on ICPR database to evaluate its performance. ICPR consists of 15 sets of images. Each set contains of 2 series of 93 images of the same person at different poses. There are 30 people in the database, having various skin colors. The pose, or head orientation are different in images. Among the images, 16 images are male and 14 images are female. The proposed approach was applied on this database images and detection rate is computed which is shown in table 3.

**Table3.** Proposed gender classification accuracy rate compared with well-known approaches on ICPR Database

| Method | Classification Accuracy (%) |
|---|---|
| **Proposed Approach** | **93.29** |
| LBP + SVM [22] | 90.65 |
| Statistical Features + Adaboost [9] | 87.65 |
| LBPH+ Statistical Features + Adaboost/SVM [25] | 91.06 |
| Patch Based + Bayesian [26] | 90.83 |

As it is shown in tables2 & 3, the proposed approach provides more classification accuracy rate than near all of other approaches. Of course, the [24] provides the best classification rate between them, but the computational complexity of [24] is more than proposed method because of computation Gabor features. In [27], the computational complexity of gabor filtering is compared with local binary patterns.

### 4.1. Run Time

The run time is an important metric to evaluate the performance of computer vision approaches. In order to evaluate the runtime, the consumed time for process a query image is considered. The runtime of the proposed approach for 90 random images is about 46.73

Preprint of published paper in:
International Journal of Signal Processing, Image Processing and Pattern Recognition
Vol. 12, No. 4 (2019), pp.1-12
http//dx.doi.org/10.33832/ijsip.2019.12.4.01
The main publisher is NADIA
seconds. It means that runtime of the proposed approach is about 519 milliseconds for each query image. So, the runtime of the proposed approach is lower than many benchmark approaches in this scope which make it usable for online applications such as smart phones.

### 4.2. Scale and rotation sensitivity

After applying the $LBP_{P,R}$ on each pixel, the occurrence probability of each label in image is computed by dividing to size of image as its shown Eq.(5). In this respect the proposed feature extracted vector (Eq.6) is probabilistic and it is not sensitive to size of human face image. As, it is mentioned in [17], the modified version of local binary patterns is rotation invariant. In Eq. (3), the uniform measure is computed by computing the transition between 0 and 1. Suppose, if the image rotates, pixel's coordination may be changed, but its neighbors value are fix because of using circular neighbors. In this respect, the number of transitions between 0 and 1, will not change, respectively uniformity measure.

## 5. CONCLUSION

An innovative approach is proposed in this paper for gender classification which can be used in smart phone applications too. In this respect, a multi-step appearance-based method is proposed based on local texture information. First, modified LBP is used to extract texture features locally. Next, an evaluated version of k-nearest neighbors is proposed which uses Tani-Moto metric as distance measure. In the result part, gender classification is done on two different databases. The classification accuracy is compared with some state-of-the-art other methods. Results show that the performance of the proposed method is higher than many other methods in the related area. Proposed approach provides some other advantages such as rotation invariant, size invariant, low runtime, and gray-scale invariant. These advantages claim the robustness of the proposed gender classification approach. Also, the proposed approach improves the performance of the smartphones by reducing RAM or CPU load complexity. Because it decreases the run time of human identification applications.

## REFERENCES

[1] *Rahman M. H., Chowdhury S., Bashar M. A.*: An automatic face detection and gender classification from color images using support vector machine, Journal of Emerging Trends in Computing and Information Science, Vol. 4, No. 1, 2013, pp. 5-11

[2] *Briones A.G., Villarrubia G., De-Paz J. F., Corchado J.M.*: A multi-agent system for the classification of gender and age form images, Computer Vision and Image Understanding, Vol. 172, 2018, pp. 98-106

[3] *Shmaglit, L., Khryashchev V.*: Gender classification of human face images based on adaptive features and support vector machines, Optical Memory and Neural Networks, Vol. 22, No. 4, 2013, pp. 228-235

[4] *Tamura S., Kawai H., Mitsumoto H.*: Male/female identification from 8×6 very low resolution face images by neural network, Pattern Recognition, Vol. 29, No. 2, 1996, pp. 331-335

[5] *Gutta S., Wechsler H., Phillips P. J.*: Gender and ethnic classification of face images. In: Third IEEE International Conference on Automatic Face and Gesture Recognition, 1998, pp. 194-199

[6] *Scalzo, F., Bebis, G., Nicolescu, M., Loss, L., Tavakkoli, A.,*: Feature Fusion Hierarchies for gender classification. In: 19[th] International Conference on Pattern Recognition, Tampa, US. 2008.

[7] *Moghaddam B., Ming-Hsuan Y.*, Learning gender with support faces, IEEE Transactions on Pattern Analysis and Machine Intelligence, Vol. 24, No. 5, 2002, pp. 707-711

[8] *Moghaddam B., Yang M. H.*: Gender classification with support vector machines. In: International Conference of Automatic Face and Gesture Recognition, 2000, pp. 306-311
10

[9] *Shakhnarovich G., Viola A. P., Moghaddam B.*: A unified learning framework for real time face detection and classification. In: International Conference on Automatic Face and Gesture Recognition, 2002

[10] *Akbari, M., Rashidi, H., Alizadeh, S. H.,*: An enhanced genetic algorithm with new operators for task scheduling in heterogeneous computing system, Engineering Applications of Artificial Intelligence, Vol. 61, 2017, pp. 35-46.

[11] *Castrillon-Santana M., Vuong Q. C.*: An analysis of automatic gender classification. In: Conference on Progress in Pattern Recognition, Image Analysis, and Applications, 2007, pp. 271–280

[12] *Buchala S., Davey N., Frank R. J., Gale T. M., Loomes M., Kanargard W.*: Gender classification of face images: The role of global and feature-based information. Lecture Notes in Computer Science, Vol. 3316, 2010, pp. 763-768

[13] *Baluja S., Rowley H. A.*: Boosting Sex Identification Performance. International Journal of Computer Vision, Vo. 71, No. 1, 2007, pp. 111-119

[14] *Makinen E., Raisamo R.*: An experimental comparison of gender classification methods, Pattern Recognition Letters, Vol. 29, No. 10, 2008, pp. 1544–1556

[15] *Pietikäinen M., Ojala T., Xu Z.*: Rotation-invariant texture classification using feature distributions. Pattern Recognition, Vol. 33, 2000, pp. 43–52

[16] *Ahonen T., Hadid A., Pietikainen M.*: Face description with local binary patterns: application to face recognition. IEEE Transactions on Pattern Analysis and Machine Intelligence, Vol. 28, No. 12, 2006, pp. 2037-2041

[17] *Ojala T., Pietikainen M., Maenpaa T.*: Multiresolution gray-scale and rotation invariant texture classification with local binary patterns, IEEE Transaction on Pattern Analysis and Machine Intelligence, Vol. 24, No. 7, 2002, pp. 971–987

[18] *Tajeripour F., Fekriershad Sh.*: Developing a novel approach for stone porosity computing using modified local binary patterns and single scale retinex, Arabian Journal for Science and Engineering, Vol. 39, No. 2, 2014, pp. 875-889

[19] *Fekriershad Sh., Tajeripour F.*: A robust approach for surface defect detection based on one dimensional local binary patterns, Indian Journal of Science and Technology, Vol. 5, No. 8, 2012, pp. 3197-3203

[20] *Takala V., Ahonen T., Pietikainen M.*: Block based methods for image retrieval using local binary patterns. Image Analysis Lecture Note in Computer Science, Vol. 3540, 2005, pp. 882-891

[21] *Caifeng S.*: Learning local binary patterns for gender classification on real world face images. Pattern Recognition Letters, Vol. 33, 2012, pp. 431-437

[22] *Lian H. C., Lu B. L.*: Multi view gender classification using local binary patterns and support vector machine. Advances in Neural Networks: Lecture Notes in Computer Science, Vol. 3972, 2006, pp. 202-209

[23] *Sang-Hyuk C.*: Taxonomy of nominal type histogram distance measures. In: American Conference on Applied mathematics, 2008, pp. 325-330

[24] *Leng X. M., Wang Y. D.*: Improving generalization for gender classification. In: 15th IEEE International Conference on Image Processing, 2008, pp. 1656–1659

[25] *Mäkinen E., Raisamo R.*: An experimental comparison of gender classification methods. Pattern Recognition Letters, Vol. 29, No. 10, 2008, pp. 1544-1556

[26] *Aghajanian J., Warrell J., Prince S. J., Rohn J. L., Baum B.*: Patch-based within object classification. In: IEEE 12th International Conference on Computer Vision, 2009, pp. 1125-1132

[27] *Tajeripour F., Kabir E., Sheikhi A.*: Fabric defect detection using modified local binary patterns, EURASIP Journal on Advances in Signal Processing, Vol. 08, 2008, pp. 1-12

[28] http://www-prima.inrialpes.fr/Pointing04/data-face.html

[29] *M. Alareqi, R. Elgouri, L. Hlou.,* High level FPGA modeling for image processing algorithms using Xilinx generator, International Journal of Computer Science and Telecommunications, Vol. 5, No. 6, 2014, pp. 1-8

[30] *Rocher R., Menard D., Scarlet P., Sentieys O.*: Analytical approach for numerical accuracy estimation of fixed points systems based on smooth operations. IEEE Transactions on Circuits and Systems, Vol. 59, No. 10, 2012, pp. 2326-2339

[31] *Tajeripour F., Saberi, M., Rezaei M., Fekri-Ershad Sh.*: Texture classification approach based on combination of random threshold vector technique and co-occurrence matrixes. In Proc. of Int. Conf. on Computer Science and Network Technology, 2011, pp. 2303-2306

[32] *Fekri-Ershad Sh., Tajeripour F.*: Color texture classification based on proposed impulse-noise resistant color local binary patterns and significant points selection algorithm. Sensor Review, Vol. 37, No. 1, 2017, pp. 33-42

[33] *Fix, E. and Hodges, J.L.* Discriminatory Analysis Nonparametric Discrimination: Consistency Properties. Technical Report 4, USAF School of Aviation Medicine, Randolph Field, Texas, USA, 1951

[34] *Nguyen, H. and Bai, L.*: Cosine Similarity Metric Learning for Face Verification. Proc. of 10[th] Asian Conf. on Computer Vision, Queenstown, New Zealand, November 8–12, 2010, pp. 709–720

[35] *Khamar, K.*: Short text classification using KNN based on distance function. Int. J. Adv. Res. Comp. Commun. Eng., Vol. 2, 2013, pp. 1916–1919

[36] *Kullback, S. and Leibler, R.A.*: On information and sufficiency. Ann. Math. Stat., Vol. 22, 1951, pp. 79–86

[37] *Golomb B. A., Lawrence D. T., Sejnowski T. J.*: SEXNET: A neural network identifies sex from human faces. Advances in Neural Information Processing Systems, 1991, pp. 572-577

[38] Wiki Dataset, https://data.vision.ee.ethz.ch/cvl/rrothe/imdb-wiki/, Accessed on 07/05/2019